\documentclass[runningheads]{llncs}

\usepackage{amscd,amssymb}
\usepackage{amsfonts}
\usepackage{graphicx,epsf}
\usepackage{enumerate,color}
\usepackage[mathcal]{euscript}
\usepackage[dvipsnames]{xcolor}
\usepackage{hyperref}
\usepackage{subfigure}
\usepackage{amsmath}
\usepackage{amssymb}
\usepackage{amsfonts}
\usepackage{dsfont}

\usepackage{algorithm}
\usepackage{algpseudocode}

\newcommand{\bx}{\mathbf{x}}
\newcommand{\bk}{\mathbf{k}}

\newcommand{\dontshow}[1]{}

\DeclareMathOperator*{\argmin}{arg\,min}
\def\kwave/{\textbf{k}-\texttt{Wave}}

\begin{document}
%\title{Iterative model correction networks for photoacoustic tomography}
%\title{Faster learned photoacoustic reconstruction using approximate k-space models}

\title{Approximate k-space models and Deep Learning for fast photoacoustic reconstruction}
\titlerunning{Approximate k-space models and Deep Learning}

\author{Andreas Hauptmann\inst{1}, Ben Cox\inst{2}, Felix Lucka\inst{1,3}, Nam Huynh\inst{2}, Marta Betcke\inst{1}, Paul Beard\inst{2}, and Simon Arridge\inst{1}}
\authorrunning{A. Hauptmann et al.}
\institute{Department of Computer Science, University College London, London, U.K. \and
Department of Medical Physics and Biomedical Engineering, University College London, London, U.K \and
Centrum Wiskunde \& Informatica, Amsterdam, Netherlands}

\maketitle
\vspace{-0.5em}
\begin{abstract}
We present a framework for accelerated iterative reconstructions using a fast and approximate forward model that is based on k-space methods for photoacoustic tomography. The approximate model introduces aliasing artefacts in the gradient information for the iterative reconstruction, but these artefacts are highly structured and we can train a CNN that can use the approximate information to perform an iterative reconstruction. We show feasibility of the method for human in-vivo measurements in a limited-view geometry. The proposed method is able to produce superior results to total variation reconstructions with a speed-up of 32 times.
\keywords{Learned image reconstruction  \and Photoacoustic tomography \and Fast Fourier methods \and Compressed sensing}
\end{abstract}

\section{Introduction}
There is increasing interest in Photoacoustic tomography (PAT) for both clinical and preclinical imaging \cite{Upputuri2016}, as it has the potential to provide molecular and functional information with high spatial resolution \cite{Beard:2011if}. For preclinical imaging it is often possible to make measurements all around the object, but for clinical imaging, PAT scanners with access to just one side of the tissue are typically required. In addition, clinical imaging typically requires high frame rates \cite{Choi2018}.  The frame rate is determined both by the time taken for the data acquisition as well as by the image reconstruction time. Compressed sensing can dramatically reduce data acquisition time, but then suitable image reconstruction approaches are required, which are typically slow due to the large number of iterations required. This paper proposes to use an approximate and fast model within a deep learning framework for PAT image reconstruction from sparse data measured using a planar scanner.

\vspace{-0.5em}

\section{Forward and inverse models}
\subsection{Photoacoustic tomography}
In PAT, a short pulse of near-infrared light is absorbed by 
chromophores in tissue. For a sufficiently short pulse, a spatially-varying pressure increase $f$ will result, which will initiate an 
ultrasound (US) pulse (\textit{photoacoustic effect}), which then propagates to the tissue surface. The measurement consists of the detected waves in space-time at the boundary of the tissue; this set of pressure time series constitutes the PA data $g$. 
%\Simon{I dropped this "two problems" comment for the sake of space. We are not talking about QPAT in this paper.}
%There are two connected inverse problems in PAT: the recovery of the %initial acoustic pressure $f$ from acoustic boundary measurements $g$, %and the estimation of chromophore distributions from measurements of %$g$ at multiple optical wavelengths. Here we consider only the former: %the linear, acoustic, part of PAT. This acoustic part is 
This acoustic propagation is
commonly modeled by the following initial value problem for the wave equation \cite{Cox2005},
\vspace{-0.5em}
\begin{equation}
(\partial_{tt} - c^2 \Delta) p(\bx,t) = 0, \quad p(\bx,t = 0) = f(\bx), \quad \partial_t p(\bx,t = 0) = 0.\label{eqn:PATfwd}
\vspace{-0.5em}
\end{equation}
The measurement of the PA signal is then modeled as a linear operator $\mathcal{M}$ acting on the pressure field $p(\bx,t)$ restricted to the boundary of the computational domain $\Omega$ and a finite time window (see \cite{Beard:2011if,LuRa13} for details on measurement systems):
\vspace{-0.5em}
\begin{equation}
g = \mathcal{M} \, p_{|\partial \Omega \times (0,T)}. \label{eqn:Measurement}
\vspace{-0.5em}
\end{equation}
Equations (\ref{eqn:PATfwd}) and (\ref{eqn:Measurement}) define a linear mapping 
\vspace{-0.5em}
\begin{equation}
Af=g,
\label{eqn:Axeqy}
\vspace{-0.5em}
\end{equation}
from initial pressure $f$ to measured pressure time series $g$, which constitutes the acoustic \emph{forward problem} in PAT. The corresponding image reconstruction step constitutes the acoustic \emph{inverse problem} to (\ref{eqn:Axeqy}). 

\vspace{-0.5em}
\subsection{Fast approximate forward and inverse models}\label{sec:kspaceModels}
\begin{figure}[h!]
\centering
\begin{picture}(280,168)
\put(0,90){\includegraphics[width=0.2\textwidth]{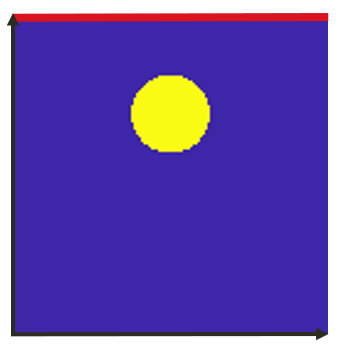}}
%\put(0,159){\linethickness{0.3mm}\textcolor{red}{\Large{\line(1,0){69.5}}}}
\put(0,-5){\includegraphics[width=0.2\textwidth]{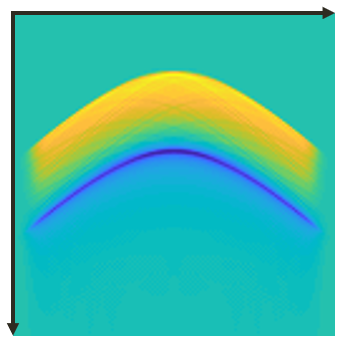}}

\put(100,90){\includegraphics[width=0.2\textwidth]{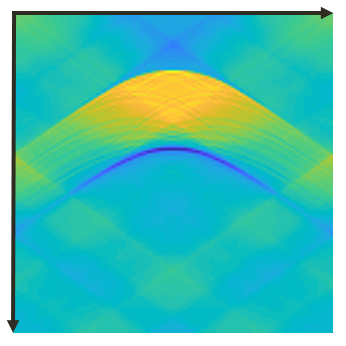}}
\put(100,-5){\includegraphics[width=0.2\textwidth]{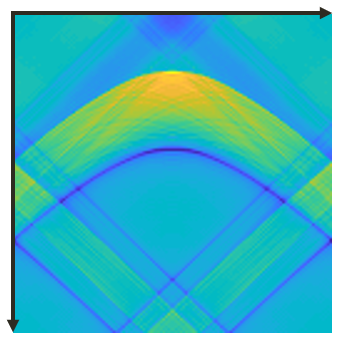}}

\put(200,90){\includegraphics[width=0.2\textwidth]{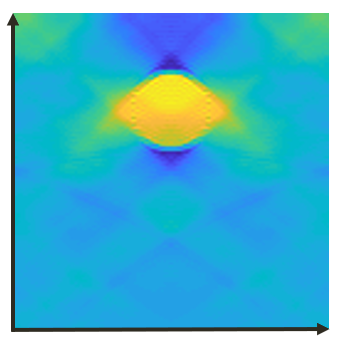}}
\put(200,-5){\includegraphics[width=0.2\textwidth]{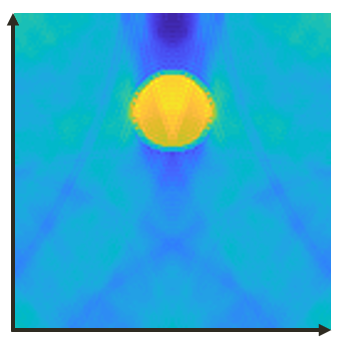}}

\put(-5,100){$z$}
\put(10,85){$x$}

\put(195,100){$z$}
\put(210,85){$x$}

\put(195,5){$z$}
\put(210,-10){$x$}

\put(-4,50){$t$}
\put(10,65){$x$}

\put(96,50){$t$}
\put(110,65){$x$}

\put(96,143){$t$}
\put(110,159){$x$}

\put(17,165){Phantom}
\put(103,173.5){k-space forward}
\put(95,165){Threshold angle: 45}
\put(200,165){k-space backward}

\put(15,72){Ideal data}
\put(103,81.5){k-space forward}
\put(95,72){Threshold angle: 80}
\put(200,72){k-space backward}

\end{picture}
\caption{\label{fig:illustrationConcept} Approximate forward model. Top left: 2D phantom with a line detector (red line).
Bottom left: ideal data. The effect of two different levels of angle thresholding of the incident waves is shown in the middle column and the resulting backprojection of the approximate data in the right column.}
\end{figure}
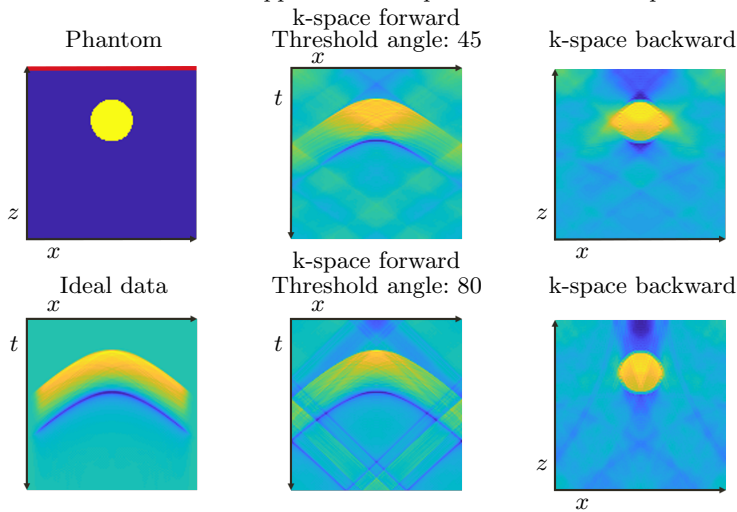

When the measurement points lie on a plane ($z=0$) outside the support of $f$, the pressure there can be related to $f$ by \cite{Cox2005}:
\vspace{-0.5em}
\begin{align}
p(x,y,t) = \frac{1}{c^2} 
\mathcal{F}_{k_x,k_y}\left\{ \left\{\mathcal{C}_{\omega}\left\{
B(k_x,k_y,\omega) \tilde{f}(k_x,k_y,\omega)
\right\}\right\}\right\},
\label{eqn:FastFwd}
\vspace{-0.5em}
\end{align}
where 
$\tilde{f}(k_x,k_y,\omega)$ is obtained from $\hat{f}(\bk)$ via the dispersion relation $(\omega/c)^2 = k_x^2+k_y^2+k_z^2$ and 
$\hat{f}(\bk) = \mathcal{F}_{\bx}\{f(\bx)\}$ is the 3D Fourier transform of $f(\bx)$. $\mathcal{C}_{\omega}$ is a cosine transform from $\omega$ to $t$, $\mathcal{F}_{k_x,k_y}$ is the 2D inverse Fourier Transform on the detector plane. The weighting factor,
\vspace{-0.5em}
\begin{align}
B(k_x,k_y,\omega) = \omega/\left(\textrm{sgn}(\omega)\sqrt{(\omega/c)^2 - k_x^2 - k_y^2}\right),
\vspace{-0.5em}
\end{align}
contains an integrable singularity which means that if Eq.\ (\ref{eqn:FastFwd}) is evaluated by discretisation on a rectangular grid, (thus enabling the application of FFT for efficient calculation), then aliasing in $p(x,y,t)$ results. An accurate model employing Eq.\ (\ref{eqn:FastFwd}) would require suitable measures to deal with the singularity, whereas evaluation using FFT leads to a \emph{fast but approximate} forward model. To control the degree of aliasing, all components of $B$ for which $k_x^2+k_y^2 > (\omega/c)^2\sin^2\theta_{\max}$ were set to zero. This is equivalent to assuming only waves arriving at angles up to $\theta_{\max}$ from normal incidence are detected. There is a trade-off: the greater the range of angles included, the greater the aliasing, as illustrated in Figure \ref{fig:illustrationConcept}.

By inverting Eq.\ \ref{eqn:FastFwd}, it can also be used as a method for mapping from the measured data $g$ to an estimate of $f$ \cite{Koestli2001}. In this case, there is no singularity to contend with, but the estimate of $f$ will suffer from limited-view artifacts \cite{Xu2004}. We will denote these two k-space methods as $A_{\mathcal{F}}$ and $A_{\mathcal{F}}^\dagger$ for the forward and backward projections, respectively.

\vspace{-1em}
%\section{Learned iterative reconstruction with approximate models}
\section{Learned reconstruction with approximate models}
In order to use an approximate forward model, such as described above, in an iterative reconstruction method, a correction must be incorporated. Here Deep Learning, specifically convolutional neural networks, offer an ideal framework to learn a correction to an approximate model. This can be done in two ways, either by learning an explicit correction of the forward model and subsequently applying an iterative scheme, or learning the correction inside a learned iterative reconstruction scheme. This study will concentrate on the second approach. 
\vspace{-0.5em}
\subsection{Learned iterative reconstruction}
Photoacoustic reconstructions from subsampled data measured over a limited detection aperture are typically computed by solving a variational problem as the minimisation of the sum of a data-fidelity term and a regularisation, $\mathcal{R}$, term enforcing certain regularities of the solution $f^*$ as%of the form
\vspace{-0.5em}
\begin{equation} \label{eqn:penaltyFunc}
f^*=\argmin_{f} \frac{1}{2}\|Af-g\|_2^2+ \alpha \mathcal{R}(f),
\vspace{-0.75em}
\end{equation}
where $\alpha>0$ is a weighting parameter. 
It has been shown in several studies \cite{Huang2013,Arridge2016,ArBeBeCoHuLuOgZh16,BoLaStGiMaBr17} that these techniques can efficiently deal with the limited view artefacts, but tend to require a larger number of iterations to converge and are additionally limited by the expressibility of the chosen regularisation term. Recently it has been shown that one can instead learn such an iterative scheme to speed up the reconstruction and additionally learn an effective regularisation for the data at hand \cite{Hammernik2018,Adler2017,Hauptmann2018}. 
This is %typically done 
achieved by formulating a simple CNN $G_{\theta_k}$, with learned parameters $\theta_k$, that computes an iterative update. Given a current iterate $f_k$, then the CNN combines $f_k$ with the gradient $\nabla d(f_k,g)$ of the fidelity term in \eqref{eqn:penaltyFunc}, such that
\vspace{-0.5em}
\begin{equation}\label{eqn:iterativeUpdate_CNN}
f_{k+1}=G_{\theta_k}(f_k,\nabla d(f_k,g)).
\vspace{-0.5em}
\end{equation}
In the following we learn each of the networks separately; i.e. starting with an initial $f_0$, we train $G_{\theta_0}$ and compute the update $f_{1}$ by \eqref{eqn:iterativeUpdate_CNN}. Then we train the subsequent networks for a set amount of iterates. This separation is done due to computational restrictions in memory and evaluation of the forward and backward projections.

\vspace{-0.5em}
\subsection{An iterative gradient network}
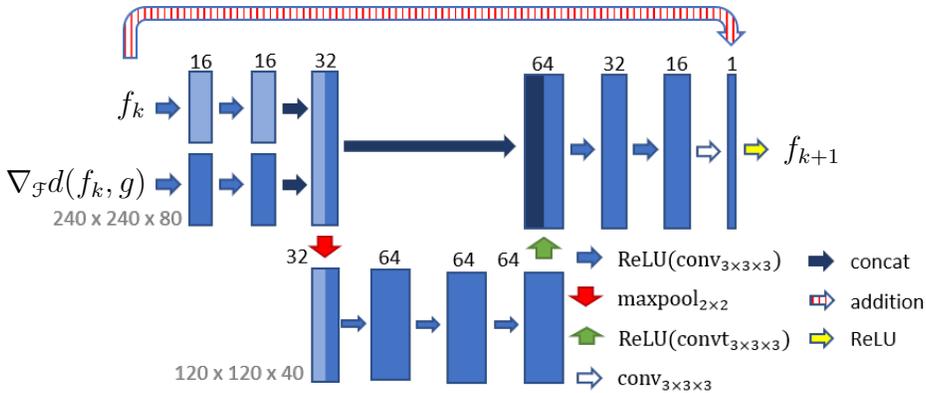
\begin{figure}[h!]
\centering
\begin{picture}(400,110)
\put(12,-20){\includegraphics[width=1\textwidth]{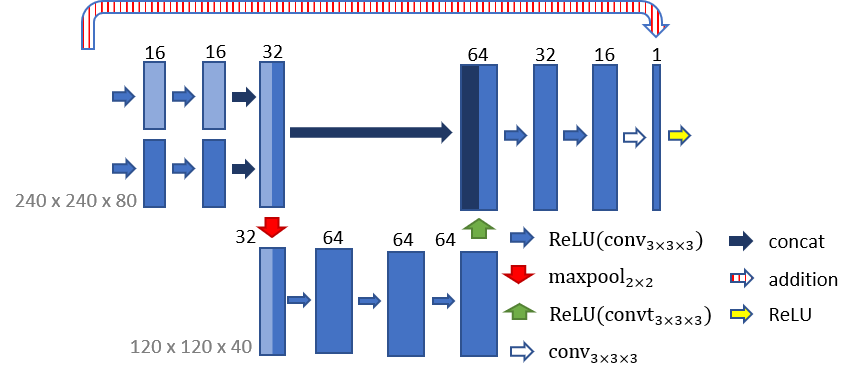}}
\put(42,88){\large $f_k$}
\put(294,71){\large $f_{k+1}$}
\put(0,58){\large $\nabla_\mathcal{F} d(f_k,g)$}
\end{picture}
\caption{\label{fig:networkStructure} Network architecture for an iterative gradient update with approximate models. Each network gets the iterate $f_k$ and the approximate gradient information $\nabla_\mathcal{F} d(f_k,g) := A_\mathcal{F}^\dagger(A_\mathcal{F}f_k-g)$ as input. The output $f_{k+1}$ is a residual update to the previous iterate. The multiscale structure is introduced to remove artefacts from the gradient.
\vspace{-0.5em}
}
\vspace{-0.5em}
\end{figure}
We propose to use an approximate model $A_\mathcal{F}$ as described in section \ref{sec:kspaceModels}. This model will be used to compute the gradient information in \eqref{eqn:iterativeUpdate_CNN}, i.e. we have $\nabla_\mathcal{F} d(f_k,g) := A_\mathcal{F}^\dagger(A_\mathcal{F}f_k-g)\approx \nabla d(f_k,g)$. By the application of the fast and approximate forward model we introduce artefacts to the gradient information, but these are highly structured, as illustrated in Figure \ref{fig:illustrationConcept}. 
Multiscale networks, such as a residual U-Net, have been proven to be efficient in detecting and removing artefacts in images \cite{Jin2017}. Thus, we believe that a multiscale network can be efficiently used to remove these artefacts. On the other hand, smaller gradient informed networks are more robust to perturbations in the measurement geometry or the imaged target, as suggested in \cite{Hauptmann2018}. 

In this work we propose to balance both approaches, by combining a deep 
gradient descent network proposed in \cite{Hauptmann2018} with a small 
mutliscale network in order to deal successfully with artefacts in the 
gradient, while still possessing the ability to generalise well with 
respect to changes in the measurement geometry. The particular network 
structure chosen for this application is illustrated in Figure 
\ref{fig:networkStructure}. The two inputs, current iterate $f_k$ and 
the approximate gradient $\nabla_\mathcal{F} d(f_k,g)$, go through two 
separate convolutional pipelines with filter size $3^3$. The results 
are then combined by concatenation and downsampled with a maxpool layer 
to a courser scale. The result of the courser scale is %then 
concatenated with the result of the two initial convolutional pipelines 
and the channel size %is 
successively reduced to one channel, which is 
added as a residual update to the input iterate $f_k$ 
%and projected to the positive numbers 
%followed by projection 
and projected onto the positive set
to produce the new iterate $f_{k+1}$.

\vspace{-0.5em}
\section{Computational results for in-vivo measurements}
\subsection{Data acquisition and preparation}
\begin{figure}[h!]
\centering
\begin{picture}(160,55)

\put(0,-10){\fbox{\includegraphics[width=0.2\textwidth]{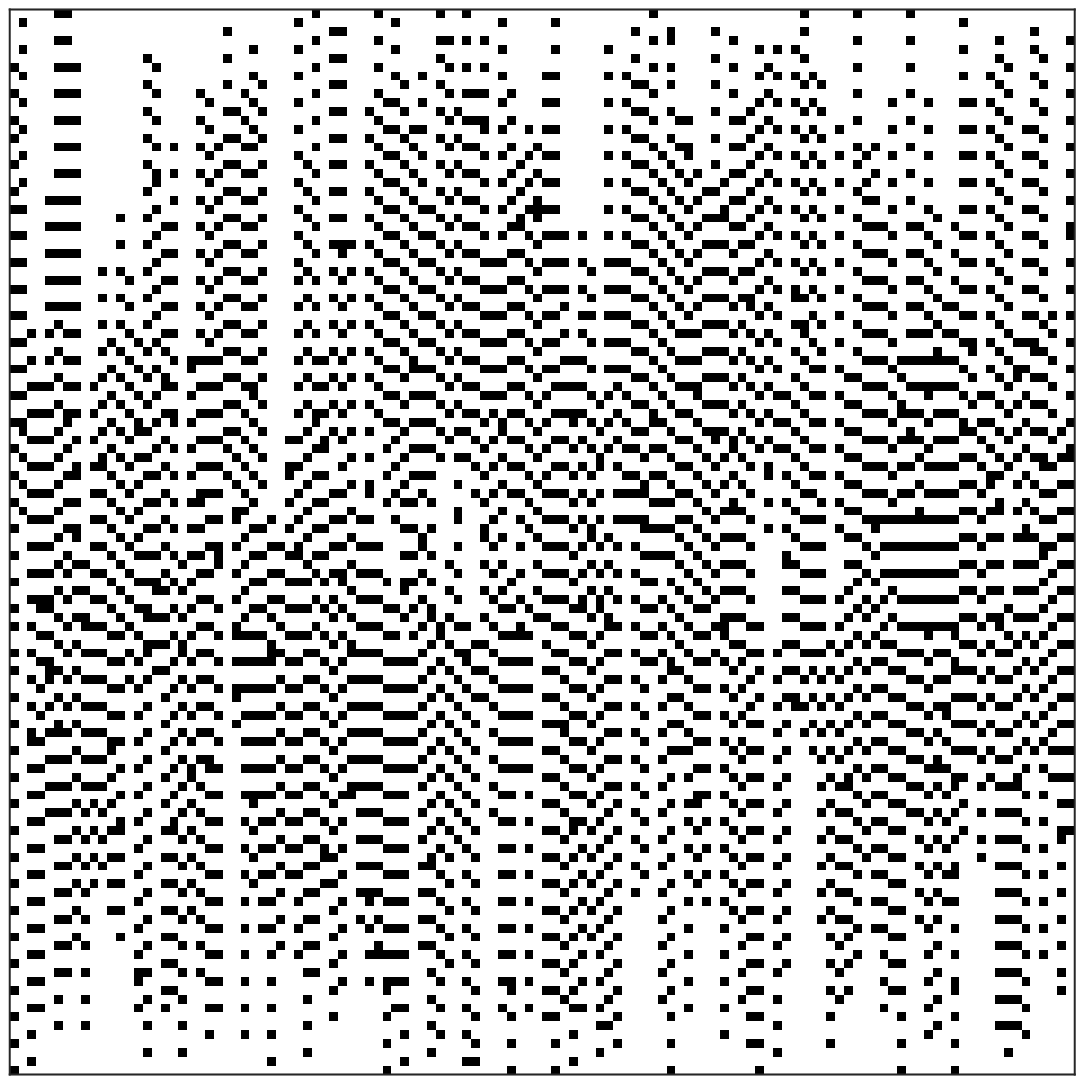}}}
\put(80,-10){\fbox{\includegraphics[width=0.2\textwidth]{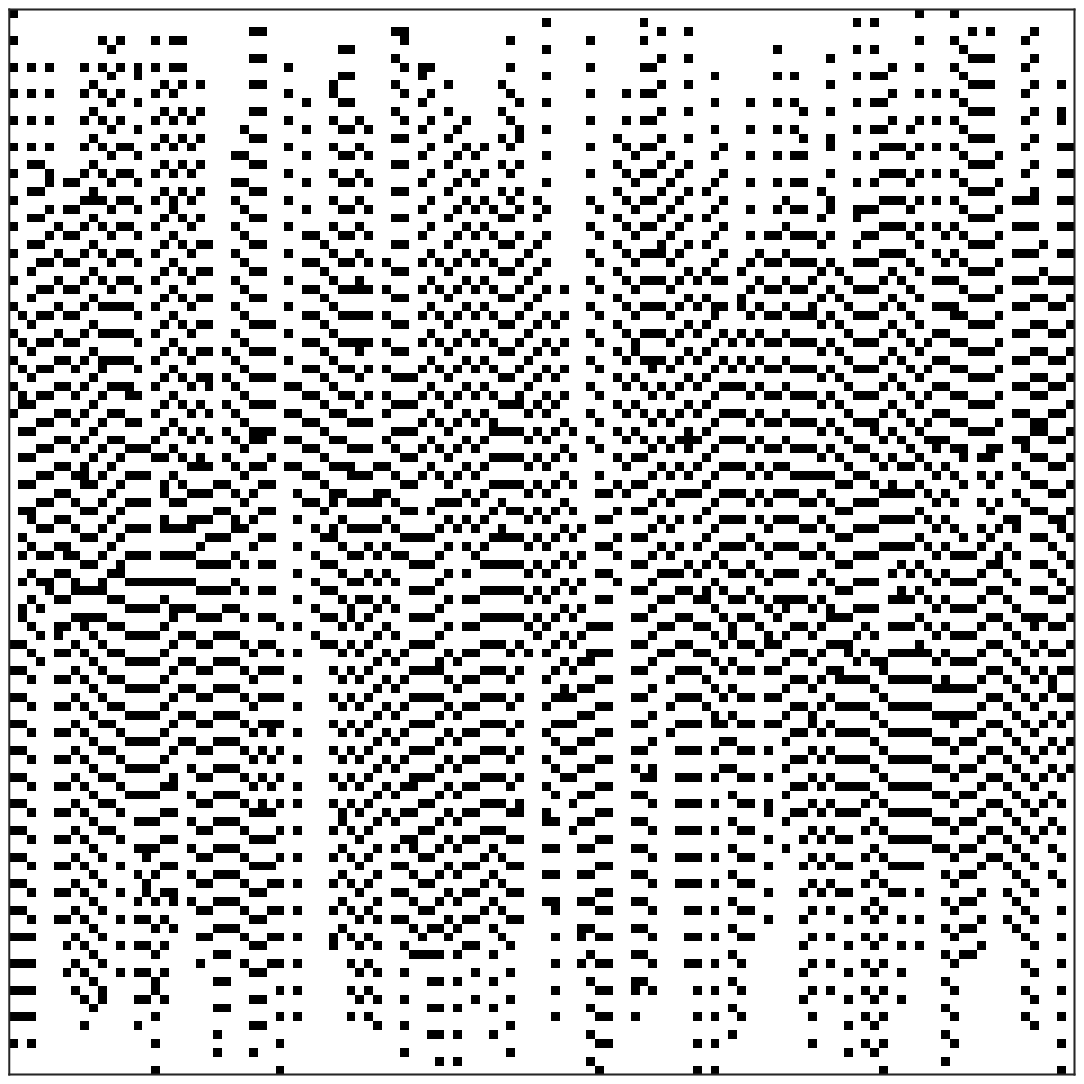}}}

\put(5,65){16 beam scanner sampling pattern}

\end{picture}
\caption{\label{fig:subPattern} Randomly generated sub-sampling pattern with the 16 beam scanner geometry and a sub-sampling factor of 4; black dots indicate interrogated points on the sensor. (Left) Pattern used for experimental sample I, (Right) pattern used for experimental sample II.
\vspace{-0.5em}}
\end{figure}

In-vivo measurements of a human subject have been taken with the planar 
sensor described in \cite{huynh2016photoacoustic}. For faster acquisition the 
scanner uses a 16 beam interrogation laser to measure the PA signal. In 
total we obtained 27 fully-sampled limited-view measurements used in 
this study. Since this is not sufficient for training an iterative 
reconstruction algorithm, we have additionally used a large dataset of 
1024 volumes of blood vessels segmented from lung CT scans as described 
in \cite{Hauptmann2018} of size $240\times 240\times 80$. We then 
simulated accurate sub-sampled limited-view photoacoustic measurement 
data of the segmented lung vessels with a sub-sampling factor of 
4 and a randomly generated 16 beam sub-sampling pattern for 
each sample, (see Figure \ref{fig:subPattern} for example 
patterns). Additionally, we have varied the sound speed in the 
simulations to be uniformly distributed in 
$[1560\text{m/s},1600\text{m/s}]$ and added normally distributed noise 
to the data with varying intensity, such that the resulting signal's 
SNR is roughly between 10 to 30. These variations have been done to 
increase robustness to variations in the measurements.

\setlength{\fboxsep}{0pt}
\vspace{-3em}
\subsection{Training of proposed network}
We have pre-trained the networks $G_{\theta_k}$ on the simulated data 
from segmented lung vessels. Given the simulated measurement $g$, the 
initial reconstruction is computed by the k-space backprojection, i.e. $f_0=A^\dagger_\mathcal{F}g$, as described in section \ref{sec:kspaceModels}. 
We have trained in total 5 iterative networks $G_{\theta_k}$ for $k=0,\dots,4$. Each network is trained in TensorFlow with the Adam 
algorithm for 30 epochs with an initial learning rate of $2\cdot 10^{-4}$ and a $\ell^2$-loss. The training of each iterate takes about 
14 hours; with initialisation and computations between iterates the whole pre-training takes a bit under 4 days on a single Titan Xp GPU.

After pre-training we have taken 25 of the in-vivo measurements and 
produced synthetically 4 times sub-sampled data with a 16 beam pattern. 
As reference reconstruction we have taken a total variation (TV) 
constrained reconstruction of the fully-sampled limited-view data. We 
have then performed an update training of the pre-trained networks with 
the 25 samples to adjust the algorithm to in-vivo artefacts not present 
in simulated data. The update training is performed for 8 epochs with a 
learning rate of $10^{-4}$ and we minimised the $\ell^2$-error to the 
reference TV reconstructions from fully-sampled limited-view data.

\vspace{-0.5em}
\subsection{Reconstructions of in-vivo measurements}
\begin{figure}[h!]
\centering
\begin{picture}(350,85)
\put(0,10){\includegraphics[height=0.225\textwidth]{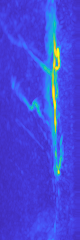}}
\put(30,-20){\includegraphics[width=0.225\textwidth]{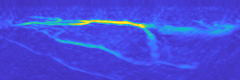}}
\put(30,10){\includegraphics[width=0.225\textwidth]{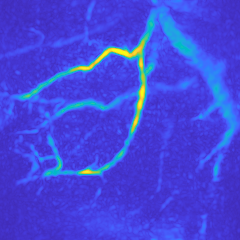}}

\put(120,10){\includegraphics[height=0.225\textwidth]{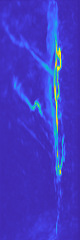}}
\put(150,-20){\includegraphics[width=0.225\textwidth]{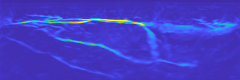}}
\put(150,10){\includegraphics[width=0.225\textwidth]{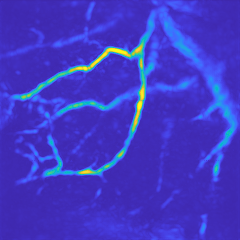}}

\put(240,10){\includegraphics[height=0.225\textwidth]{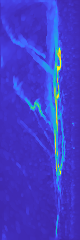}}
\put(270,-20){\includegraphics[width=0.225\textwidth]{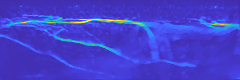}}
\put(270,10){\includegraphics[width=0.225\textwidth]{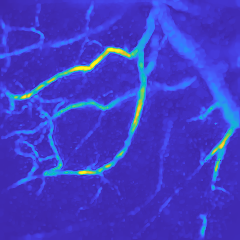}}

 \put(15,95){Initial backprojection}
 \put(135,95){FF-PAT, 5 iterations}
 \put(250,95){fully-sampled reference}

\end{picture}
\caption{\label{fig:reconSample1} Sample I: reconstruction of in-vivo measurements from 4$\times$ undersampled 16-beam pattern (maximum intensity projections). PSNR in comparison to the reference from fully-sampled limited-view data: backprojection 33.5672, FF-PAT 42.1749.}
\end{figure}
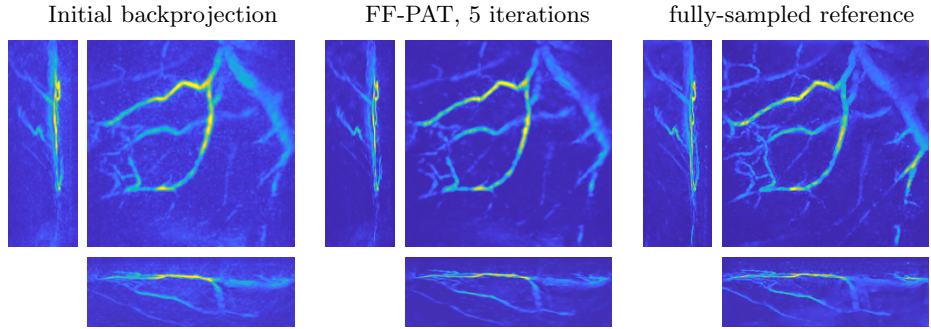

\begin{figure}[h!]
\centering

\begin{picture}(350,90)
\put(0,10){\includegraphics[height=0.225\textwidth]{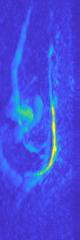}}
\put(30,-20){\includegraphics[width=0.225\textwidth]{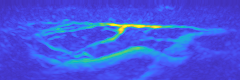}}
\put(30,10){\includegraphics[width=0.225\textwidth]{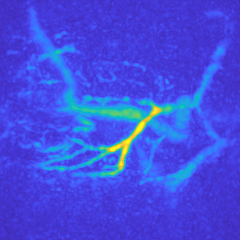}}

\put(120,10){\includegraphics[height=0.225\textwidth]{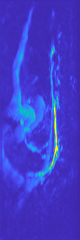}}
\put(150,-20){\includegraphics[width=0.225\textwidth]{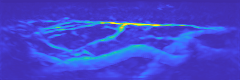}}
\put(150,10){\includegraphics[width=0.225\textwidth]{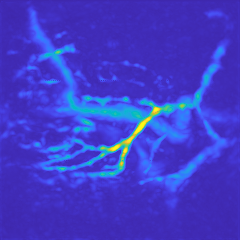}}

\put(240,10){\includegraphics[height=0.225\textwidth]{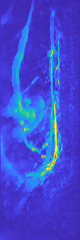}}
\put(270,-20){\includegraphics[width=0.225\textwidth]{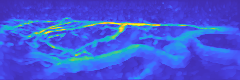}}
\put(270,10){\includegraphics[width=0.225\textwidth]{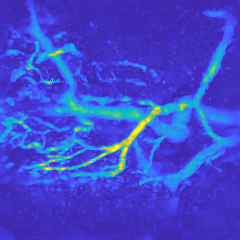}}

 \put(15,95){Initial backprojection}
 \put(135,95){FF-PAT, 5 iterations}
 \put(250,95){fully-sampled reference}

% \put(-10,20){\includegraphics[height=0.25\textwidth]{imagesReal/fastPAT_16Beam_4x_sample_4_iter_0_mIY.png}}
% \put(25,-15){\includegraphics[width=0.25\textwidth]{imagesReal/fastPAT_16Beam_4x_sample_4_iter_0_mIX.png}}
% \put(25,20){\includegraphics[width=0.25\textwidth]{imagesReal/fastPAT_16Beam_4x_sample_4_iter_0_mIZ.png}}

% \put(120,20){\includegraphics[height=0.25\textwidth]{imagesReal/fastPAT_16Beam_4x_sample_4_iter_5_mIY.png}}
% \put(155,-15){\includegraphics[width=0.25\textwidth]{imagesReal/fastPAT_16Beam_4x_sample_4_iter_5_mIX.png}}
% \put(155,20){\includegraphics[width=0.25\textwidth]{imagesReal/fastPAT_16Beam_4x_sample_4_iter_5_mIZ.png}}

% \put(250,20){\includegraphics[height=0.25\textwidth]{imagesReal/fastPAT_16Beam_4x_sample_4_ref_mIY.png}}
% \put(285,-15){\includegraphics[width=0.25\textwidth]{imagesReal/fastPAT_16Beam_4x_sample_4_ref_mIX.png}}
% \put(285,20){\includegraphics[width=0.25\textwidth]{imagesReal/fastPAT_16Beam_4x_sample_4_ref_mIZ.png}}

% \put(100,90){\includegraphics[width=0.2\textwidth]{images/forwProj_fast_45.png}}
% \put(100,00){\includegraphics[width=0.2\textwidth]{images/forwProj_fast_80.png}}

% \put(200,90){\includegraphics[width=0.2\textwidth]{images/backProj_fast_45.png}}
% \put(200,00){\includegraphics[width=0.2\textwidth]{images/backProj_fast_80.png}}

%  \put(25,112){Initial backprojection}
%  \put(154,112){FF-PAT, 5 iterations}
%  \put(282,112){fully-sampled reference}

\end{picture}
\caption{\label{fig:reconSample2} Sample II: reconstruction of in-vivo measurements from 4$\times$ undersampled 16 beam pattern (maximum intensity projections). PSNR in comparison to the reference from fully-sampled limited-view data: backprojection 34.4372, FF-PAT 42.0388.}
\vspace{-0.5em}
\end{figure}

\begin{figure}[h!]
\centering
\begin{picture}(250,105)
\put(0,20){\includegraphics[height=0.225\textwidth]{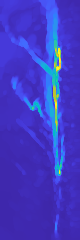}}
\put(30,-10){\includegraphics[width=0.225\textwidth]{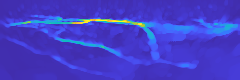}}
\put(30,20){\includegraphics[width=0.225\textwidth]{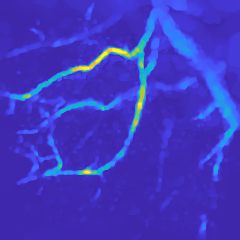}}

\put(120,20){\includegraphics[height=0.225\textwidth]{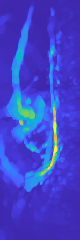}}
\put(150,-10){\includegraphics[width=0.225\textwidth]{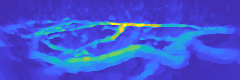}}
\put(150,20){\includegraphics[width=0.225\textwidth]{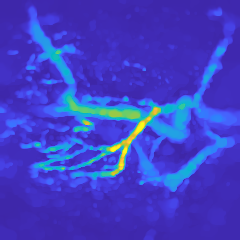}}

\put(50,102){Sample I}
\put(170,102){Sample II}
%\put(90,125){TV reconstructions}

\end{picture}
\caption{\label{fig:reconTV} TV reconstructions (20 iterations, maximum intensity projections) of in-vivo measurements from 4$\times$ undersampled 16-beam pattern. PSNR in comparison to the reference from fully-sampled limited-view data: Sample I 41.1576, Sample II 42.1391.}
\vspace{-0.5em}
\end{figure}

The reconstruction with the trained network is performed on 2 samples of in-vivo limited-view measurements with 4 times sub-sampling, the corresponding sub-sampling pattern is shown in Figure \ref{fig:subPattern}. The resulting reconstructions for both samples are shown in Figure \ref{fig:reconSample1} and \ref{fig:reconSample2}. Evaluation of the projections take each 1.6 seconds and of the network 0.45 seconds, hence one iterate takes a bit less than 4 seconds. The total computation time for 5 iterates with initialisation is about 20 seconds on a single Titan Xp GPU.  
For comparison we have computed TV reconstructions of the same sub-sampled data for both test cases. The regularisation parameter was chosen, such that PSNR to the reference reconstruction is maximised. The resulting reconstructions are shown in Figure \ref{fig:reconTV} and take approximately 11 minutes.

% \begin{table}[h] 
% {\centering
% \scriptsize
%   \caption{Measures for in-vivo experiment: in comparison to reference TV reconstruction from fully-sampled limited-view data}
%     \begin{tabular}{l|c|c|c|| c l|c|c|c} 
%      %\multicolumn{5}{c}{\sc Reconstructed Average Values}\\
%     %\hline
% Sample I&{\sc Init. } &{\sc fast PAT} & {\sc TV (20 Iter)} & \ \ &Sample II&{\sc Init. } &{\sc fast PAT} & {\sc TV (20 Iter)} \\
%     \hline
%     \hline 
%     {\sc PSNR}  	  & 26.4961 & 39.9393 & 39.3251  & 	&  & 23.8594 & 36.5423 & 36.7947  \\
%     %\sc SSIM} (??)   & 0.3292 & 0.8908 &  0.8880  		&  &  & 0.2107 & 0.8488 & 0.8495 \\
%     {\sc Time (sec.)}  &  & & 702 & & &  &  &  661
%     \end{tabular}%
%   \label{table:inVivoErrors} }
% \end{table}% 

\vspace{-3em}
\subsection{Discussion}
In both cases, the image quality of the Fast Forward PAT (FF-PAT) reconstructions is clearly improved with respect to the initial backprojection. Even though we have used approximate projection operators, the results suggest that the proposed network generalises well and incorporates the approximate gradient in a useful manner. In comparison to the TV reconstruction, FF-PAT is competitive with respect to PSNR computed in comparison to the reference reconstructions: higher for Sample I and similar for Sample II. In terms of visual quality, the FF-PAT reconstructions can be considered superior due to strong blocky artefacts present in the TV reconstructions, especially in the background where small details are present (compare in Sample II). Furthermore, reconstruction times are reduced by a factor of 32. In comparison to learned iterative reconstructions with the accurate model, see \cite{Hauptmann2018}, image quality is competitive with a speed-up of FF-PAT by factor 8.

\section{Conclusions}
Iterative reconstructions are necessary in restricted measurement 
geometries to successively negate limited-view artefacts. This involves 
the repeated evaluation of forward and backward projections, which can 
be costly in high-resolution and 3D. We have successfully shown that 
one can use approximate models instead in a learned iterative 
reconstruction algorithm, where the network also learns to negate 
approximation artefacts in the gradient. We achieve a speed-up of up to 32 compared to established TV reconstructions and providing 
superior reconstructions. While this study applies for planar sensors 
in PAT, the framework can be extended to different measurement 
geometries and possibly other modalities.

%\dontshow{
\section*{Acknowledgment}
Support of NVIDIA Corporation with one Titan Xp GPU.
AH is supported from the Wellcome-EPSRC project NS/A000027/1. FL is supported from EPSRC project EP/K009745/1 and the Netherlands Organization for Scientific Research (NWO), project nr. 613.009.106/2383.
%}
%\vfill

\vspace{-0.5em}
\bibliographystyle{unsrt}
\bibliography{literature,PAT,PATCS}
%\bibliography{literature,PAT}

\end{document}